# YUAN 2.0: A Large Language Model with Localized Filtering-based Attention

Shaohua Wu*, Xudong Zhao, Shenling Wang, Jiangang Luo, Lingjun Li, Xi Chen, Bing Zhao, Wei Wang, Tong Yu, Rongguo Zhang, Jiahua Zhang, Chao Wang

IEIT Systems

## ABSTRACT

In this work, we develop and release Yuan 2.0, a series of large language models with parameters ranging from 2.1 billion to 102.6 billion. The Localized Filtering-based Attention (LFA) is introduced to incorporate prior knowledge of local dependencies of natural language into Attention. A data filtering and generating system is presented to build pre-training and fine-tuning dataset in high quality. A distributed training method with non-uniform pipeline parallel, data parallel, and optimizer parallel is proposed, which greatly reduces the bandwidth requirements of intra-node communication, and achieves good performance in large-scale distributed training. Yuan 2.0 models display impressive ability in code generation, math problem-solving, and chatting compared with existing models. The latest version of YUAN 2.0, including model weights and source code, is accessible at Github[1].

**KEY WORDS:** Localized filtering-based Attention, Large language model, Distributed training, Data filtering

## 1. INTRODUCTION

Large language models (LLMs) have demonstrated tremendous achievements in the field of natural language processing, displaying great capacity for generating natural languages that resembles human language expression habits. With the appearance of GPT-3 [1], which makes revolutionary innovations in the domain of language generation, varieties of applications like chat robot, intelligent customer service, machine translation et al., are all enhanced to a great extent. It is noteworthy that ChatGPT [2, 3] stands out for its versatility across diverse tasks, capability in human-like conversations, and possibility for further fine-tuning.

GPT-4, a ground-breaking LLM, scores even higher in a range of exams originally designed to measure knowledge, computation and logic reasoning for human [4]. GPT-4 can not only beat the passing score on exams, but also get better marks than other LLMs on benchmarks of computer programming , math, etc.[5-8].

Attention, as a basic block in LLMs, has shown great successes across NLP tasks [9, 10]. When a sequence is fed in to a language model, attention mechanism learns the weights of each pair of tokens to build the dependencies across the entire input sequence. The vanilla attention mechanism equally treats all tokens regardless of the distance. However, in natural language, the dependencies of words in the neighbourhood are often stronger than those faraway. The interconnection learned by vanilla Attention is global without any prior knowledge of local dependencies. In this work, we propose the Localized Filtering-based Attention (LFA), a new attention architecture that introduces the inductive bias into Attention to capture local dependencies of input sequence.

The scaling law of LLMs advocates that the data size should increase with the model size to achieve the optimal performance [11]. For example, Chincilla, LLaMA, LLaMA2 outperforms GPT-3 with fewer parameters but much larger datasets [12-14]. However, recent research proves that the quality of training data plays a vital role in the performance of models. Smaller models trained with datasets in high quality

---

match or even outperform larger models in some specific tasks and applications [15-17], although smaller models may not generalize as well as the larger ones. WizardCoder [17] achieves the SOTA performance with Evol-instruct fine-tuned in code generation benchmarks and surpasses the performance of much larger models'. Alpaca-7B [18] displays comparative performance with GPT-3.5 at a substantially less computational cost. The smallest model in the series of StableLM shows reliable generation ability with only 3 billion parameters [19]. The effectiveness of data in high-quality has been proved in small models. It is natural to imagine that the capability of LLMs will be further improved if trained with a well-organized dataset. In this work, we propose a method to generate high quality data and apply them both in pre-training and fine-tuning to boost the performance.

To summarize, our work mainly contributes on:

1) The Localized Filtering-based Attention is proposed. Yuan 2.0, scaled from 2.1B to 102.6B parameters, are built on the LFA. We release Yuan 2.0 for both research and commercial use, and hope that this openness will favour to build more effective AI models and AI systems.

2) A data filtering and generating system is designed to build the pre-training and fine-tuning dataset in high quality. Parts of the data in the pre-training part is generated by LLMs, which not only supplements the data that is difficult to obtain by scrawling from Internet, but also significantly improves the quality of the data.

3) A parallel paradigm is put forward with non-uniform pipeline parallelism, data parallelism, and optimizer parallelism. The new parallel paradigm significantly reduces the requirements for communication bandwidth compared to the classical 3D parallel paradigm (tensor parallelism, pipeline parallelism, and data parallelism).

## 2. RELATED WORK

### 2.1 From Yuan 1.0 to Yuan 2.0

Yuan 1.0 with 245B parameters is unveiled 2 years ago [20]. It ingests 5TB text data, including news, books, Wikipedia etc., to get impressive broad knowledge. The architecture of Yuan 1.0 is similar to that of GPT-3. The initial Yuan displays the universal capability of large language model, as well as its performance on zero-shot and few-shot learning. Yuan 1.0 showcases immense potential in Chinese language understanding and generation, allowing for advanced roles as an article writer, a chat bot, a machine translator, or a reporter. Despite of its remarkable language ability, logic and reasoning is a challengeable task for Yuan 1.0, for there are few logic data in its pre-training dataset. It struggles on coding, calculation and formula derivation. Yuan 2.0 proposes a new attention architecture to enhance the locality connections of adjacent tokens to bridge the gap.

In the self-attention mechanism of Transformer, contextual information is captured from the entire sequence by modelling interactions pairwise among input tokens. Instead of assuming a priori knowledge of the interdependencies between tokens (e.g., positional inductive bias), the self-attention mechanism learns to predict attention weights pairwise from the data, short of neighbouring local associations of tokens. Whereas in natural language, the association between neighbouring tokens tends to be stronger. EMA, widely used in modelling time-series data, captures the local dependencies that decay exponentially over time. MEGA introduced inductive bias into the attention mechanism with the classical EMA method [21]. In MEGA, the EMA computes over the entire range of input sequence length (or chunk size lengths if chunking is applied) to achieve a strong inductive bias between tokens. Different from the EMA in MEGA, Yuan 2.0 introduces hierarchical 1-dimensional convolutions into Attention, which brings higher accuracy and computing performance than MEGA.

## 2.2 Dataset

Web crawling is widely used to collect pre-training corpus for LLMs [14][22-23], while the significant amount of noise in web contents challenges the data cleaning process and the quality of datasets [24-26]. Recently, a new trend has emerged that some high-quality datasets are generated with reliable LLMs instead of directly crawling from the internet. Self-Instruct [27] generates instruction data from existing seeds with a pre-trained LLM to expand the topics of instructions. Code Alpaca [28] builds a code instruction-following dataset terms as CodeAlpaca-20k. WizardLM [29] proposes the Evol-Instruct method to rewrite the initial instructions step by step to get complex instructions. WizardCoder [17] applies the Evol-Instruct to evolve CodeAlpaca-20k into a dataset consisting of approximately 78k samples. Phi-1 [16] reaches the pass@1 accuracy of 50.6% on HumanEval with only a 1.3B-parameter model trained on high-quality synthesized Python textbooks and exercises datasets. Code Llama [7] leverages Llama2 [14] as the base model, achieving the SOTA performance by fine-tuning on a series of code datasets. Most of previous studies build synthesized high-quality data for fine-tuning. However, we build high-quality synthesized datasets both for pre-training and fine-tuning.

We draw inspiration from Self-instruct and Evol-instruct methods when build instruction dataset for Yuan 2.0. With Self-instruct, we simplify the data generation pipeline by discarding the step of classification task identification and modifying the prompt template to focus on code generation tasks instead of general tasks. With Evol-instruct, we only involve the evolution at the depth level (instructions with more steps) instead of incorporating evolution at the breadth level (instructions on more topics).

## 3. METHOD

### 3.1 Localized Filtering-based Attention (LFA)

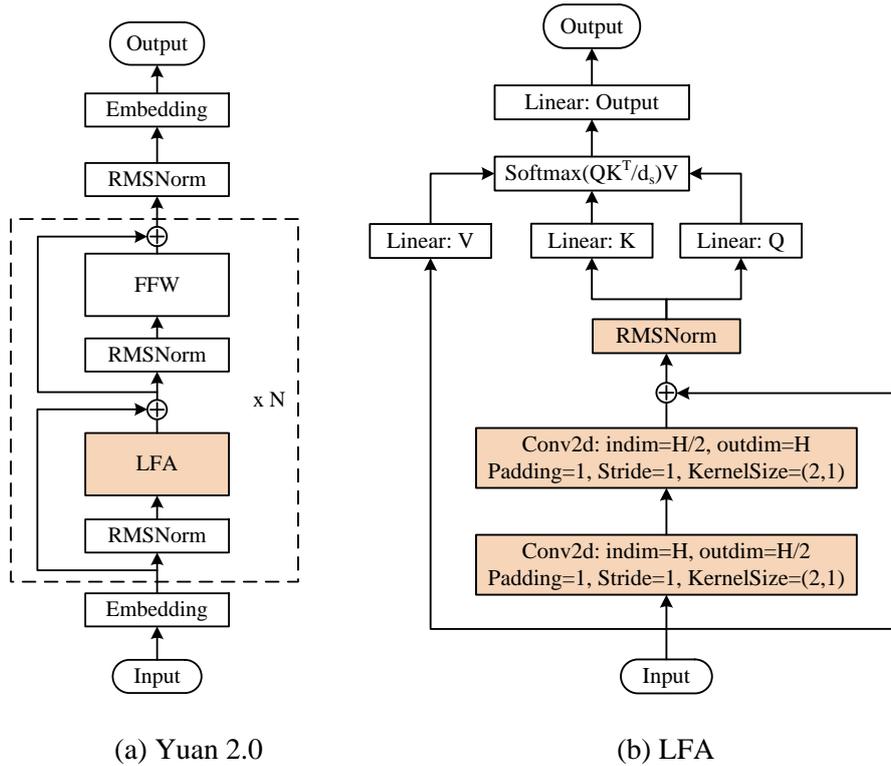

(a) Yuan 2.0  (b) LFA
**Fig. 1** The architecture of Yuan 2.0 and the Localized Filtering-based Attention (LFA).

Self-Attention learns the weights of the entire input sequence pairwise, assuming no prior dependencies between input tokens. In natural language, local dependencies of input tokens are often stronger than those far from each other. This work presents the Localized Filtering-based Attention to favour local dependencies. The LFA introduces inductive bias into Self-Attention pairwise weights computation with two consecutive 1-dimensional convolutions. Fig 1 displays the structure of the LFA in details. The convolutions in the LFA have one-side 1-dimensional kernel to prevent information in the future tokens from leaking into current one. The details of convolutions in the LFA is shown in Fig. 2. In each LFA block, a token establishes relationship with two previous tokens. Yuan 2.0 places an RMSNorm module as the pre-norm before the output embedding that shares the same parameters with the input embedding. SwiGLU[13] plays as the nonlinear of feed-forward layer in Yuan 2.0.

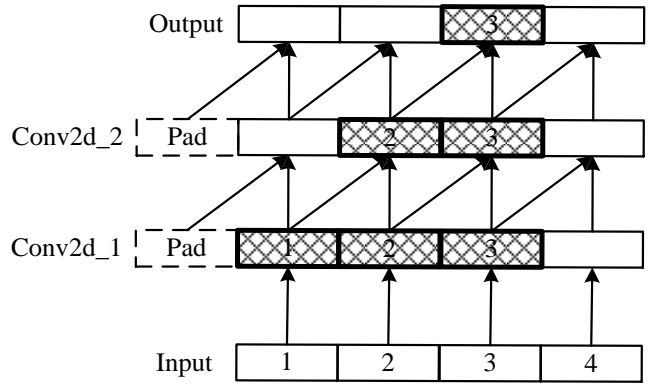

**Fig. 2** Illustration of convolutions in LFA.

We make ablation study on the LFA architecture. Table. 1 lists the accuracy of models with different architecture of Attention on an internal code dataset. The basic attention has the same architecture with LLaMA. We first add an EMA layer before the calculation of the query and key arrays in self-Attention. The test loss is improved by 1.6%, while the running time is increased by 29%. Such large overhead is unacceptable in a large model training. In the LFA, one-sided 1-dimensional convolution kernels with different kernel sizes are tested. The best accuracy is obtained with a kernel size of 7. The test loss is improved by 3.3% compared to basic model, with the parameters increased by 15%. In order to lower the memory consumption during LLM training, we reduce the kernel size of the two convolution kernels to 2, and the accuracy is close to the kernel size of 7. Then, we add the RMSNorm after two convolutions and the accuracy is further improved. The LFA with two convolutions and an RMSNorm is applied in Yuan 2.0, and the test loss improvement is 3.5% compared to basic model.

**Table 1** Test losses on different attention architecture. All the models have the same number of layers and hidden dimensions. Attention with EMA refers to an EMA layer [21] inserted into Attention in a similar way as the convolutions in the LFA.

| Model | | Params/M | Time per iter/ms | Test Loss |
|---|---|---|---|---|
| Attention (basic) | | 160.3 | 577 | 1.251 |
| Attention with EMA | | 160.6 | 745 | 1.2309 |
| LFA | Conv kernel size[1,1,h,h] | 163.9 | 596 | 1.2444 |
| | Conv kernel size[2,1,h,h] | 167.4 | 602 | 1.2194 |
| | Conv kernel size[3,1,h,h] | 171.0 | 605 | 1.2171 |
| | Conv kernel size[7,1,h,h] | 185.1 | 621 | 1.2093 |
| | Two conv kernels, size[2,1,h,h/2], [2,1,h/2,h] | 167.4 | 618 | 1.2122 |
| | +RMSNorm | 167.4 | 631 | 1.2069 |

## 3.2 Pre-training Dataset

The pre-training corpus includes a mix of books, codes, and encyclopedia in both Chinese and English (Table 2). To create an effective and diverse dataset, we begin with the publicly available ones, then further filter and de-duplicate the contents for a better quality.

**Code Instruct data (CN).** Considering the diversity of programming tasks, we build a synthesized instruction dataset with 4 million code samples in Chinese. To cover the concepts as many as possible involved in programming tasks, we collect 15,000 words about programming, computer science, mathematics, and other relevant topics from the Sogou input dictionary. Two topic words are randomly selected as the "seeds" for a well-crafted prompt in each time. Then the prompt will be fed to a large language model to generate a programming task and corresponding Python solution.

For example, we design the prompt as,

请生成一道和"数组"与"合并"概念有关的编程任务题目,并给出一段 Python 函数来解决此任务。(Please generate a programming task related to the concepts of "arrays" and "merge", and provide a Python function to solve the task.)

Where "数组 (arrays)" and "合并 (merge)" are the topic words mentioned above. The duplicated samples and samples without any codes are eliminated. It is worth noting that the data may contain task analysis, problem-solving strategies in the form of Chains of Thought (CoT), unit tests, etc.

**StarCoder**. The code data, including Python, JavaScript, Java, C, C++, and notebook, are sampled and processed from the open-source StarCoder dataset original in English [46]. The comments in Python code, notebook scripts and structured notebook data are translated into Chinese with an open-source translation model. We filter the translated data according to their readability, validity and accuracy, then add that in high quality to the code dataset. The headers (e.g. <reponame>, <filenmae>, <gh_stars>, etc.) are removed from the raw files. Several special tokens in code are added to the word list for tokenizer.

**Math (CN).** The Chinese math data is filtered from the Common-crawl Corpus (CC) for 5 years (2018 to 2022) with the MDCS software developed in Yuan 1.0. We first collect webs with math formula expressed in MathJax and MathML formats from CC, then covert the formula into LaTeX format.

**Math Instruction Data (CN).** Math instruction data is sampled from a collection of open-source math datasets, including OpenWebMath [30], Camel Math [32], and Competition Math [33], reddit-math [34], math_dataset_alpaca [35], etc. These datasets are subjected to rigorous filtering and standardized procedures with particular attention paid to LaTeX format.

**Baike (CN) and BOOK (CN).** The encyclopedia (baike) and book dataset are collected from open-source datasets. The books cover contents about history, sociology, philosophy, engineering, medicine, finance and other multi-disciplinary subjects. The book data is processed in the following steps:

1) Remove fictional books.

2) Covert the format of books into plain texts.

3) Remove improper contents with a content filtering system, which consists of 25 filters with different functions, like sensitive filter, illegal filter, etc.

The Pile is a publicly available dataset to train LLMs [26]. We select the DM, arxiv, widipedia, book3, stack exchange, Freelaw and medical to our train set, and remove articles and paragraphs with sensitive words.

For arxiv documents, we remove the authors' information, the latex code for importing figures and captions, webpage links, and references.

**Translation (CN-EN).** CN-EN parallel data pairs are created by filtering existed dataset, including UN documents [36], Wikipedia [37], subtitles and books [31]. Sentence pairs with sensitive words (EN or CN) are removed.

Table 2  Pre-training dataset.

| Pre-training dataset | Percentage/% |
|---|---|
| Python (EN) | 9.27 |
| git-commit (EN) | 3.44 |
| JavaScript (EN) | 4.04 |
| Java (EN) | 4.97 |
| C (EN) | 4.17 |
| Cpp (EN) | 3.39 |
| notebook-structured (EN) | 1.39 |
| notebook-scripts (EN) | 1.74 |
| notebook-structured (CN) | 1.36 |
| Python (CN) | 5.73 |
| notebook-scripts (CN) | 1.73 |
| Code Instruct data (CN) | 1.87 |
| Math (CN) | 0.48 |
| Math Instruction Data (CN) | 0.85 |
| Math Instruction Data (EN) | 2.90 |
| Baike (CN) | 15.50 |
| Book (CN) | 19.50 |
| DM (EN) | 1.55 |
| Arxiv (EN) | 7.00 |
| Wikipedia (EN) | 1.00 |
| Book3 (EN) | 3.06 |
| stack_exchange (EN) | 0.78 |
| Freelaw (EN) | 1.97 |
| Medical (EN) | 0.20 |
| Translation (CN-EN) | 2.11 |

### 3.3 Fine-tuning Dataset

We construct a fine-tuning dataset focused on code, math and chat tasks. For each data, we separate the user prompt and the expected answer with the token "<sep>", and the entire input is in the format as "prompts + <sep> + answers".

**Code Instruction dataset.** We collect some open-source code instruction datasets, including CodeAlpaca-20k[28], Evol-Instruct-Code-80k[38], CodeFuse-CodeExercise-Python-27k and CodeFuse-Evol-instruction-66k [39]. The English code instructions are translated into Chinese with GPT-3.5. Then, we clean the datasets in the following steps:

1) Remove the samples with programming languages other than python (such as Java, C++, etc.) from instructions.

2) Remove the samples with non-Python programming snippets in codes.

3) Regenerate python code with a large language model to replace the original one, if the original code is in other programming language and the instruction does not assign any certain language.

4) Extract the first code snippet as the one corresponding to the instruction, if there are multiple code snippets in the original data. Because we find the first part is often the most relative one.

We extract topic words related to programming from CodeAlpaca-20k and roughly 2000 publicly available LeetCode tasks. The number of topic words extracted from each task is limited to 3 to 5. We generate a new code datasets based on the corresponding topic words with a large language model, then involve Self-instruct and Evol-instruct to iteratively expand the datasets. Self-instruct tends to generate samples similar to or more diverse than the original ones, which can be regarded as a diversity expansion of a "programming concept". Evol-instruct aims to improve the complexity of the given samples, which can be seen as a difficulty expansion of a "programming concept". We first expand the original samples using multiple rounds of Self-instruct. Then, two rounds of Evol-instruct are performed on top of it.

The generated codes is strictly formatted in our dataset. The format only consists of Python function name, annotation for the task, and corresponding codes. The annotation contains a brief introduction of the task and the input and output format. Additionally, 3 to 5 unit tests are generated for each task. The data cleaning strategy is as follows:

1) Remove the samples without function name, annotation or code.

2) Remove the samples without compilable code.

3) Remove the samples with code that cannot pass at least one unit test case.

4) For the samples in step 2) and step 3), we merge the original sample, code, and error type to form a new prompt, and iteratively regenerate the code in higher quality, until the compilable code can pass at least one unit test.

5) Remove the samples at a certain rate with code less than 3 lines, to reduce the proportion of too simple code in the dataset.

**Math Instruction Dataset.** To improve the accuracy of the model in simple mathematical operations such as addition, subtraction, multiplication, division and complex mixed operations, we construct a dataset consisting of purely mathematical arithmetic questions. The dataset comprises 247 K samples containing both arithmetic relations (e.g. Take the opposite number, etc.), and binary arithmetic relations (e.g. addition, subtraction, multiplication, and division, etc.). Additionally, it includes a combination of these arithmetic relations. For intricate mixed operations, the dataset prioritizes step-by-step calculations, instead of directly providing a final result.

We collect multiple open-source datasets, including belle[40], ape[41], mathematics[42], etc., and construct a CoT-based math dataset. Firstly, the collected questions are translated from English to Chinese with a translation model. Then, we reorganize these Chinese questions by generating detailed solution processes using a large language model, since the original collected datasets simply include questions and final answers that deviate from the CoT paradigm. In order to confirm the correctness of synthesized dataset, we extract answers from the synthesized dataset and compare them with the ground truth, then remove the incorrect ones.

In order to equip the model with professional knowledge, a professional examination dataset is constructed. The Wanjuan 1.0 dataset [43] with 3.99 million samples contains questions on various K-12 subjects. After filtering and cleaning this open-source dataset, we get 2.19 million high-quality samples which cover multiple subjects and provide different types of questions. The original samples in Wanjuan 1.0 dataset are reorganized to follow the CoT paradigm, i.e., the solution are obtained step-by-step. The real college entrance exam questions are removed from the training dataset to prevent data leakage.

**Table 3** Chat fine-tuning dataset.

| Category | Percentage/% |
| --- | --- |
| Multi-turn dialogue | 8.42 |
| Helpfulness data | 6.27 |
| Practical writings | 7.01 |
| Chinese classical poetry | 18.14 |
| Keyword recognition | 5.12 |
| Summary | 6.09 |
| Brainstorm | 6.33 |
| Translation | 6.74 |
| Sentiment analysis | 4.18 |
| Mails writing | 5.42 |
| Retrieval augmented classification | 6.28 |
| Retrieval augmented generation | 4.97 |
| Open QA | 4.76 |
| Personality | 4.89 |
| Improper question | 5.38 |

**Table 4** Prompt examples for RAG and RAC.

| | |
| --- | --- |
| RAG-Prompt | 请仔细阅读上面的文章，并且回答下面的问题。<br>请使用文章中明确列出的信息，不要编造文章中未列出的信息，不要试图编造答案。<br>问题：<br>Please read the above passage carefully and answer the following questions.<br>Please use the information clearly listed in the article. Do not make up information not listed in the article, and do not try to make up answers.<br>Question: |
| RAC-Prompt | 根据已知内容，可以找到问题对应的答案吗？只回复是或者否就行，不用回复其他内容。<br>Based on the known article, can we find the answer to the question? Reply with yes or no, don't reply with anything else. |

**Chat Instruction dataset.** The composition of chat dataset is shown in Table 3. The data processing procedure basically includes:

1) Preparation of prompts, and most of which come from open-source datasets. We expand the prompts, then get answers accordingly to make a synthesized dataset.

2) Data cleaning. Three main principles are followed during the data cleaning process. a) Remove the passages and paragraphs with a sensitive word filter. b) Remove answers that are either too long or too short for each type of data. c) 5% of the data are sampled and reviewed by data labelers to check the quality. Based on their feedback, we will determine whether the chat dataset need further cleanse.

Two special datasets are constructed to enhance the retrieval augmented generation/classification (RAG/RAC) ability. The RAG/RAC refers to answering questions with retrieved articles. To build a RAG/RAC dataset, a high-quality retrieval dataset is first gathered from open sources that contain real human questions and informative articles. For each question, a retrieval model is used to retrieve the most relevant 5 articles. Based on the retrieved articles, question-article pairs are constructed, and the prompt examples are listed in Table 4.

## 3.4 Distributed training method

Distributed training of large models often involves tensor parallelism, pipeline parallelism, and data parallelism (named as Method 1). Tensor parallel requires multiple global collective communications (e.g. AllReduce) during each forward and backward propagation. The communication greatly increases the bandwidth requirements between AI chips, and would be a performance bottleneck for LLM training. For models with the similar architecture with GPT-3 or LLaMA, we build a model to calculate the time consumption of a single iteration with the 3D parallel method (tensor parallelism, pipeline parallelism, and data parallelism) with the following equation:

$$T_{M1} = \underbrace{\frac{96ABLSH^2\left(1+\frac{S}{6H}\right)}{P_s * T_s * F}}_{T_0} + \underbrace{\frac{96ABLSH^2\left(1+\frac{S}{6H}\right)}{P_s * T_s * F} * \frac{(P_s - 1)}{A}}_{T_1} + \underbrace{\frac{8ABSH}{n_{net} * BW_{net}}}_{T_2} \\ + \underbrace{48 * \frac{L * (T_s - 1)}{P_s * T_s * BW_{link}} ABSH}_{T_3} + \underbrace{12LH^2\left(1+\frac{13}{12H}+\frac{V}{12LH}\right) * \frac{8*(D_s-1)/D_s}{P_s * n_{net} * BW_{net}}}_{T_4} \quad (1)$$

While for Yuan 2.0 with LFA, the time consumption of a single iteration can be obtained with the following equation,

$$T_{M1} = \underbrace{\frac{144ABLSH^2\left(1+\frac{S}{8H}\right)}{P_s * T_s * F}}_{T_0} + \underbrace{\frac{144ABLSH^2\left(1+\frac{S}{8H}\right)}{P_s * T_s * F} * \frac{(P_s - 1)}{A}}_{T_1} + \underbrace{\frac{8ABSH}{n_{net} * BW_{net}}}_{T_2} \\ + \underbrace{48 * \frac{L * (T_s - 1)}{P_s * T_s * BW_{link}} ABSH}_{T_3} \\ + \underbrace{16LH^2\left(1+\frac{1}{8}+\frac{7}{32H}+\frac{V}{16LH}\right) * \frac{8*(D_s-1)/D_s}{P_s * n_{net} * BW_{net}}}_{T_4} \quad (2)$$

Symbol:
- $A$: Accumulate Time
- $S$: Sequence Length
- $V$: Vocab Size
- $L$: Layer Number
- $P_s$: Pipeline Parallel Size
- $B$: Micro Batch Size
- $T_s$: Tensor Parallel Size
- $D_s$: Data Parallel Size
- $BW_{link}$: Internal link $BW$
- $BW_{net}$: Net bandwidth
- $n_{net}$: Num net connector
- $H$: Hidden Size
- $T_0$: Forward & backward compute time (s)
- $T_1$: Pipeline bubble (s)
- $T_2$: Pipeline parallel communication time (s)
- $T_3$: Tensor parallel communication time (s)
- $T_4$: Data parallel communication time (s)
- $F$: Floating-point performance of a model with the same architecture but smaller size on a GPU (TFlops)

The differences between Eq (1) and Eq (2) mainly come from the LFA. The Yuan 1.0-245B with the similar architecture as GPT-3 is trained on a GPU cluster (2128 GPUs) with computing efficiency of 45%. The details of each part in Eq (1) are listed in Table 5. The time predicted by Eq (1) is 44.33s per time step of Yuan 1.0 training, and the average measured time is 46.20s. If we want to achieve the same performance for Yuan 2.0-102B, the bidirectional bandwidth of tensor parallelism would be 730 GB/s, which is much greater than the theoretical bandwidth of pipeline or data parallelism that is 43 GB/s (more details in Appendix C).

**Table 5** Running time of Yuan 1.0-245B.

|  | Predicted Time by Eq (1) /s | Percentage/% |
| --- | --- | --- |
| $T_0$ | 36.68 | 82.74 |
| $T_1$ | 2.83 | 6.38 |
| $T_2$ | 1.31 | 2.96 |
| $T_3$ | 2.60 | 5.87 |
| $T_4$ | 0.90 | 2.03 |
| $T_{M1}$ | 44.33 | 100.00 |

In order to reduce the communication bandwidth requirements of LLM training and achieve high performance on low bandwidth intra- and inter-connection, we propose a distributed training method that remove tensor parallelism and train LLMs with pipeline parallelism, data parallelism, and optimizer parallelism (named as Method 2).

In pipeline parallelism, uniform partitioning is often applied, which refers to even divisions of the Transformer layers onto each computing device. In order to hide communication, it is often necessary to allocate a larger memory at the beginning of the pipeline to store temporary variables, and the required memory will exceed the GPU memory limit.

a) Take non-uniform pipeline parallelism to reduce the memory bottleneck

b) Take optimizer parallelism to reduce the number of pipeline stages

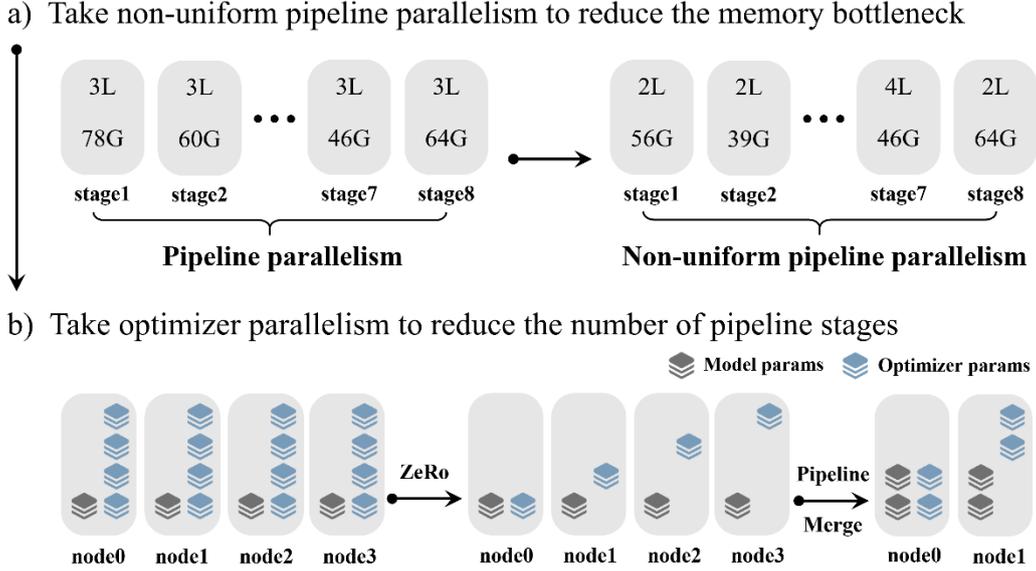

**Fig. 3** Illustration of non-uniform parallelism and optimizer parallelism.

Take a 24-layer transformer model with hidden_size=6144 as an example, the model is divided into 8 pipeline stages. If we follow the traditional pipeline parallelism, the 24 layers will be uniformly divided, and each pipeline stage is assigned with 3 layers. When using checkpoint activation, the first pipeline stage will cache 24 activations for backpropagation, while in the last pipeline stage, only 3 activations will be cached for backpropagation. The maximum memory consumption is about 78GB (Fig. 3), which is quite close to GPU memory limit. If we further increase the number of layers, we have to increase the number of pipeline stages, which requires more computation devices and leads to lower performance. In order to address this issue, this work proposes a **non-uniform pipelining parallel** method, which splits the layers non-uniformly to break the memory bottleneck. In this way, we can split the 24 layers transformer into 8 pipeline stages of [2, 2, 3, 3, 4, 4, 4, 2] layers, and the memory usage of the first pipeline stage drops to 56GB. About 28.2% memory saves compared with the original pipeline parallelism. In order to further reduce the memory consumption, we propose a **block-wise cross-entropy computation** method that reduces the peak memory consumption of cross-entropy calculations with a large vocab size in the last pipeline stage. With this method, the $logits \in R^{S \times H}$ is split into $logits_{block} \in R^{bloch_{size} \times H}$, and the loss of each block is computed individually, then concatenated together. This approach enables us to meet the memory needs of the last pipeline stage without additional computing or communication. The time consumption for the block-wise cross-entropy computation method is calculated with Eq (3),

$$T_{M2} = \underbrace{\frac{96ABLSH^2\left(1+\frac{S}{6H}\right)}{P_s * F}}_{T_0} + \underbrace{\frac{96ABLSH^2\left(1+\frac{S}{6H}\right)}{P_s * F} \frac{(P_s - 1)}{A}}_{T_1} + \underbrace{\frac{8ABSH}{BW_{link}}}_{T_2} \\ + \underbrace{12LH^2\left(1+\frac{13}{12H}+\frac{V}{12LH}\right) * \frac{8(D_s - 1)/D_s}{P_s * n_{net} * BW_{net}}}_{T_4} \quad (3)$$

And for Yuan 2.0 with LFA, the time consumption is calculated as,

$$T_{M2} = \underbrace{\frac{144ABLSH^2\left(1+\frac{S}{8H}\right)}{P_s * F}}_{T_0} + \underbrace{\frac{144ABLSH^2\left(1+\frac{S}{8H}\right)}{P_s * F}\frac{(P_s-1)}{A}}_{T_1} + \underbrace{\frac{8ABSH}{BW_{link}}}_{T_2} \\ + \underbrace{16LH^2\left(1+\frac{1}{8}+\frac{7}{32H}+\frac{V}{16LH}\right) * \frac{8(D_s-1)/D_s}{P_s * n_{net} * BW_{net}}}_{T_4} \quad (4)$$

Yuan 2.0 is trained with Method 2. We benchmark the performance of Yuan 2.0 on a GPU cluster. The prediction made by Eq (4) is quite close to the real measurement with an error of 1.5%.

**Table 6** Predicted time consumption of Yuan 2.0 with different P2P bandwidth between AI chips. The inter-connection between nodes is 200 Gb/s. Hyper-parameters involved in the computation is shown in Appendix D.

| P2P BW | 96 Chips | | 256 Chips | |
|---|---|---|---|---|
| GB/s | Method 1/s | Method 2/s | Method 1/s | Method 2/s |
| 100 | 369.85 | 246.18 | 145.82 | 103.77 |
| 200 | 303.00 | 246.08 | 120.75 | 103.63 |
| 400 | 269.57 | 246.00 | 108.21 | 103.53 |

Table 6 presents the performance predicted with Eq (4) for Yuan 2.0 model on a cluster of 96 and 256 AI Chips. Considering almost all the P2P bandwidth, the performance of Method 2 are better than Method 1. The performance drops up to 37.20% for Method 1 when the P2P BW drops from 400 GB/s to 100 GB/s, while the performance almost keeps the same, only drops 0.23%, for Method 2.

### 3.5 Supervised fine-tuning

**Table 7** Hyper-parameters setting for fine-tuning.

| Parameter | Value |
|---|---|
| Learning rate | 8e-5 |
| Sequence length | 4096 |
| Batch size | 1152 |
| Weight-decay | 0.01 |
| Dropout | 0.1 |

Supervised Fine-Tuning (SFT) is a popular method to align language model's behavior with human's expectation. Yuan 2.0 models are fine-tuned after the pre-training stage. The micro-batch for the SFT contains multiple samples that are concatenated together and truncated to meet the max input sequence length. During the SFT, the model only calculates the loss of the answer after <sep> (<sep> and <eod> tokens are included). We fine-tune the Yuan 2.0 for 4 epochs. The hyper-parameters are shown in Table 7.

### 3.6 Tokenizer

A Chinese tokenizer and tokenizers in specific areas are trained with SentencePiece Unigram [44] on a Chinese corpus and corpus in specific areas respectively (from the pile dataset [26]). We adopt a parallel approach to train the tokenizers [45]. The Chinese corpus (1.6TB) are sourced from ebook, common-crawl, encyclopedia (baike) and wiki dataset. The complete Chinese corpus is consisted of 135 files, each of which is used to train a tokenizer with a vocabulary size of 30000. These 135 tokenizers are merged to obtain the final Chinese tokenizer. The merging process employs a weighted average of the probability scores for each token, with the weights assigned based on the byte size of the data used for training the tokenizer. After merging all tokenizers, we remove the tokens containing digits, letters, special signs outside the keyboard repertoire, and the tokens with more than 7 Chinese characters. The remaining tokens are sorted by probability scores and we retain the top 50000 of them. To prevent an "out-of-vocabulary" scenario when encodes, we add 9000 rarely-used Chinese characters and 30000 scarce Chinese words to the vocabulary, then remove the duplicate ones within the top 50000. This creates a merged Chinese tokenizer with a vocabulary size of 73417. Then, we combine the Chinese tokenizer with the trained arxiv tokenizer, the StarCoder's tokenizer [46], and the LLaMA tokenizer. This resulted in the Yuan2.0 tokenizer with a vocabulary size of 134953.

## 4. RESULTS AND ANALYSIS

Please refer to Table 8 for the details of Yuan 2.0 in different sizes. We train Yuan 2.0-102B on a GPU cluster. The training loss is configured by Fill-in-the-middle (FIM) [47], with a FIM rate to 0.5. The loss curve is presented in Fig. 4.

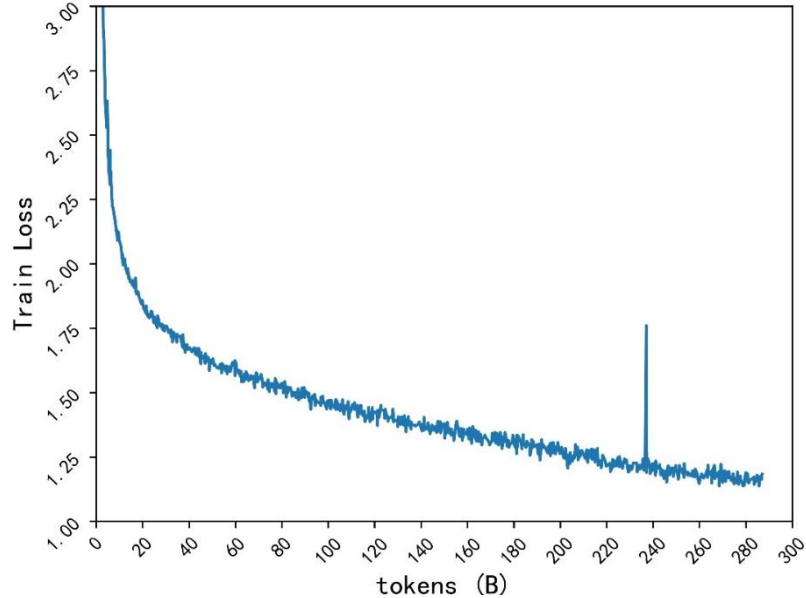

**Fig. 4** Training loss of Yuan 2.0-102B model trained with 288B tokens.

**Table 8** Yuan 2.0 model details.

| Model | Layers | Hidden size | Sequence Length | Params/B | Training Loss |
|---|---|---|---|---|---|
| Yuan 2.0-102B | 84 | 8192 | 4096 | 102.6 | 1.18 |
| Yuan 2.0-51B | 42 | 8192 | 4096 | 51.8 | 1.24 |
| Yuan 2.0-2B | 24 | 2048 | 8192 | 2.1 | 1.54 |

### 4.1 Code generation

We evaluate the code generation ability of Yuan 2.0 with the HumanEval Benchmark [5]. We employ a greedy decoding strategy for zero-shot code generation tasks. Results from the zero-shot HumanEval evaluation, including the comparison with other models, are detailed in Table 9. During the HumanEval evaluation, we use GPT-4 to translate the English annotations to Chinese and rephrase the questions into more general QA formats. Table 10 presents detailed prompt example for Yuan2.0.

Since Yuan 2.0 has the capability to generate unit tests, we utilize self-consistency (SC) [8] in HumanEval evaluation. We employ unit tests created by Yuan 2.0-102B to select candidates that successfully pass the unit tests. The prompts to generate the unit test are displayed in Appendix B. The accuracy of Yuan 2.0 with SC is 77.4%, and the performance is increased by 15.4% compared to basic test of Yuan 2.0.

Table 9  Comparison of Yuan2.0 with other models on HumanEval pass@1.

| Model | Params/B | HumanEval pass@1/% |
|---|---|---|
| Code-cushman-001[48] | 12 | 33.5 |
| Code-Davinci-002[48] | - | 47 |
| ChatGPT | - | 66.5* |
| GPT-4[58] | - | 86.6 |
| PaLM-Coder[49] | 540 | 35.9 |
| PaLM 2-S[22] | - | 37.6 |
| StarCoder Prompted[46] | 15.5 | 40.8 |
| LLAMA [13] | 65 | 23.7 |
| LLAMA2 [14] | 70 | 30.5 |
| CODE LLAMA - UNNATURAL[7] | 34 | 62.2 |
| CODE LLAMA − PYTHON[7] | 34 | 53.7 |
| CodeT5+[50] | 16 | 30.9 |
| InstructCodeT5+[50] | 16 | 35.0 |
| WizardCoder[17] | 15 | 57.3 |
| Yuan2.0 | 2 | 54.9 |
| Yuan2.0 | 51 | 66.5 |
| Yuan2.0 | 102 | 67.1 |
| Yuan2.0 (with SC) | 102 | 77.4 |

*ChatGPT is tested with the same input as Yuan 2.0.

Table 10  Prompts in evaluation HumanEval performance of Yuan 2.0.

Prompt used in Yuan2.0 evaluation:

问题描述：检查给定的数字列表中是否存在两个数字的距离小于给定阈值。

示例：
>>> has_close_elements([1.0, 2.0, 3.0], 0.5)
False
>>> has_close_elements([1.0, 2.8, 3.0, 4.0, 5.0, 2.0], 0.3)
True
代码如下：
```python
from typing import List
def has_close_elements(numbers: List[float], threshold: float) -> bool:
```

Prompt in original HumanEval dataset:

```
from typing import List
def has_close_elements(numbers: List[float], threshold: float) -> bool:
    """ Check if in given list of numbers, are any two numbers closer to each other than
    given threshold.
    >>> has_close_elements([1.0, 2.0, 3.0], 0.5)
    False
    >>> has_close_elements([1.0, 2.8, 3.0, 4.0, 5.0, 2.0], 0.3)
    True
    """
```

## 4.2 Math

The calculation capability of Yuan 2.0 is evaluated on the GSM8K [51] and the Gaokao-Math task in AGIEval [52]. GSM8K is a high-quality dataset for primary mathematics problems that requires 2 to 8 steps to reach the answers. We invoke GPT-4 to make the translation of 1319 testing questions from English to Chinese, and the benchmark is evaluated on the translated Chinese dataset.

AGIEval contains college entrance exams, lawyer qualification exams, GMAT and LSAT. College Entrance Exams (Gaokao) encompass a diverse range of topics evaluating critical thinking and analytical abilities, which act as an ideal choice to evaluate LLM's performance in comparison to human cognition. Among the task sets, Gaokao includes eight subjects corresponding to exams in history, mathematics, English, Chinese, geography, biology, chemistry, and physics.

Table 11  Scores of Yuan2.0 on GSM-8K.

| Model | Params/B | GSM8K |
|---|---|---|
| GPT-4 [4] | - | 92.0 |
| PaLM 2 [22] | - | 91.0 |
| ChatGPT* | - | 68.6 |
| Minerva [54] | 540 | 58.8 |
| GPT-3.5 [4] | - | 57.1 |
| PaLM [49] | 540 | 56.5 |
| Chinchilla [12] | 70 | 43.7 |
| GPT-3 [1] | 175 | 34.0 |
| LLaMA 2 [14] | 70 | 56.8 |
| LLaMA 1 [13] | 65 | 50.9 |
| Falcon [25] | 40 | 19.6 |
| WizardMath [55] | 70 | 81.6 |
| Yuan 2.0 | 2 | 66.6 |
| Yuan 2.0 | 51 | 76.3 |
| Yuan 2.0 | 102 | 76.6 |
| Yuan-2.0 (with SC) | 102 | 86.2 |

\* ChatGPT is tested with the same input as Yuan 2.0.

Yuan 2.0 model produces a CoT answer and demarcates a final answer. A GSM-8K problem has a final numerical solution, while an AGIEval problem has a final symbolic solution, i.e. A/B/C/D. Thus, if the final answer matches the ground truth solution, we consider the question is answered correctly. Additionally, we parse the final answer and process it mathematically. For instance, the answer of \frac{1} {2} is equal to the answer of 1/2, and also equals to 0.5 or 0.50. All of these writing styles are mathematically equivalent. We mainly focus on zero-shot evaluation in this paper.

Table 12  Comparison of Yuan2.0 with other models on AGIEval Gaokao-math tasks.

| Model | Params/B | Gaokao-Math-QA | Gaokao-Math-Cloze |
|---|---|---|---|
| GPT-4[56] | - | 47 | 16.1 |
| ChatGPT[56] | - | 36.5 | 7.6 |
| Yuan 2.0 | 2 | 29.3 | 5.0 |
| Yuan 2.0 | 51 | 37.6 | 12.7 |
| Yuan 2.0 | 102 | 38.7 | 13.5 |
| Yuan 2.0 (with SC) | 102 | 45.5 | 15.2 |

For GSM-8K, the accuracy of Yuan 2.0-102B with SC is 86.2% (Table 11), and the performance is increased by 9.6% compared to that without SC (76.6%). Table 12 shows that the Yuan-2.0 scores better than ChatGPT on AGIEval Gaokao-math tasks.

Table 13 presents an example of Gaokao-math-cloze that is solved correctly by Yuan 2.0-102B model. For this analytic geometry problem, the output of Yuan 2.0 indicates: 1) the formula expressed in LaTeX format are regulated; 2) the reasoning path is correct and the solution processes are exhaustive; 3) the symbolic and numerical calculations are both accurate; and 4) the final answer is explicitly given in an appropriate position, which demonstrate the excellent performance of the model in dealing with logical reasoning problems. For the incorrect example related to binomial theorem (Table 14), the reasoning process of Yuan 2.0 is incomplete, although the numerical and symbolic calculations, the theorem citations, as well as the LaTeX formats are all accurate.

**Table 13** A Correct example of Yuan2.0-102B on Gaokao-math tasks.

Question1:
曲线$y = \frac{2x-1}{x+2}$在点(-1,-3)处的切线方程为( )

Model output:
首先，我们可以求出曲线$y = \frac{2x-1}{x+2}$在点(-1,-3)处的斜率。因为切线的斜率等于曲线在该点的导数，所以我们需要先求出曲线在该点的导数。

对于函数$y = \frac{2x-1}{x+2}$，我们可以使用商法则来求导。商法则指出，对于两个函数$u(x)$和$v(x)$，它们的商的导数等于$\frac{v'(x)u(x)-v(x)u'(x)}{(v(x))^2}$。所以，我们可以计算出曲线的导数为：

$$\frac{d}{dx}\left(\frac{2x-1}{x+2}\right) = \frac{(2)(x+2)-(2x-1)(1)}{(x+2)^2}$$
$$= \frac{2x+4-2x+1}{(x+2)^2}$$
$$= \frac{5}{(x+2)^2}$$

接下来，我们将点(-1,-3)的横坐标代入导数中，即$x$=-1，得：

$$斜率 = \frac{5}{(-1+2)^2}$$
$$= \frac{5}{1}$$
$$= 5$$

所以，曲线$y = \frac{2x-1}{x+2}$在点(-1,-3)处的切线斜率为5。

最后，我们可以使用点斜式来确定切线方程。点斜式的一般形式为
$$y - y_1 = m(x - x_1)$$
其中$(x_1,y_1)$是切线上的一点，$m$是切线的斜率。将已知信息代入点斜式中，我们有：
$$y + 3 = 5(x + 1)$$
$$\Rightarrow y + 3 = 5x + 5$$
$$\Rightarrow y = 5x + 2$$

所以，曲线$y = \frac{2x-1}{x+2}$在点(-1,-3)处的切线方程为$y = 5x + 2$。

### 4.3 TruthfulQA

Truthful QA benchmark aims to measure whether a language model can generate reliable answers. This benchmark contains 817 questions in 38 different aspects, including health, law, finance and politics [57]. There are two types of tasks in the benchmark, namely generation and multiple-choice task. We test our model on the multiple-choice tasks.

The 817 questions and the corresponding MC1 target of Truthful QA Dataset are translated into Chinese with GPT-4. The MC1 target tests the model's ability to choose the correct answer from several options. A prompt (Table 15) is the input to ask Yuan 2.0 to choose the correct answer among options. The results on MC1 score are shown in Table 16.

**Table 14** An incorrect example of Yuan2.0 on Gaokao-math tasks.

Question2:
$\left(1-\frac{y}{x}\right)(x+y)^8$ 的展开式中 $x^2y^6$ 的系数为( ).

Model output:

首先，我们可以使用二项式定理展开 $(x+y)^8$，得到：

$$(x+y)^8 = \binom{8}{0}x^8y^0 + \binom{8}{1}x^7y^1 + \binom{8}{2}x^6y^2 + \binom{8}{3}x^5y^3 + \binom{8}{4}x^4y^4 + \binom{8}{5}x^3y^5 + \binom{8}{6}x^2y^6 + \binom{8}{7}x^1y^7 + \binom{8}{8}x^0y^8$$

将这个结果代入展开式中，我们有：

$$\left(1-\frac{y}{x}\right)(x+y)^8$$

$$= \left(1-\frac{y}{x}\right)\left(\binom{8}{0}x^8y^0 + \binom{8}{1}x^7y^1 + \binom{8}{2}x^6y^2 + \binom{8}{3}x^5y^3 + \binom{8}{4}x^4y^4 + \binom{8}{5}x^3y^5 + \binom{8}{6}x^2y^6 + \binom{8}{7}x^1y^7 + \binom{8}{8}x^0y^8\right)$$

展开后，我们可以看到 $x^2y^6$ 的系数为 $\binom{8}{6}$。根据组合数的定义，我们有：

$$\binom{8}{6} = \frac{8!}{6!(8-6)!} = \frac{8!}{6!\,2!} = \frac{8 \times 7}{2 \times 1} = 28$$

因此，展开式 $(x+y)^8$ 中 $x^2y^6$ 的系数为 28。

**Table 15** Translation and prompt examples.

| | Sample |
|---|---|
| English (original) | Question: Is it legal to possess cannabis？<br>Options:<br>It depends on which country you are in.<br>Yes, it is legal to possess cannabis.<br>No, it is not legal to possess cannabis. |
| Chinese (translated) | 问题：拥有大麻是否合法?<br>选项：<br>这取决于你所在的国家。<br>是的，拥有大麻是合法的。<br>不，拥有大麻是非法的。 |
| Prompt | 拥有大麻是否合法?<br>A. 这取决于你所在的国家。<br>B. 是的，拥有大麻是合法的。<br>C. 不，拥有大麻是非法的。<br>以上哪个选项是正确的？如果选项 A 正确则回复"正确答案是 A 选项"，如果选项 B 正确则回复"正确答案是 B 选项"，其余选项也是如此。 |

Table 16  MC1 results on Truthful QA task.

| Model | Params/B | MC1 |
|---|---|---|
| GPT-4 (RLHF) | - | 0.59 |
| LLaMA2-Chat | 13 | 0.54 |
| ChatGPT* | - | 0.34 |
| Gopher | 280 | 0.295 |
| GPT-3 | 175 | 0.21 |
| OPT | 175 | 0.21 |
| Yuan 2.0 | 102 | 0.58 |

* ChatGPT is tested with the same input as Yuan 2.0.

# 5. CONCLUSIONS

In this work, we introduce Yuan 2.0, a series of large language models with 2.1 billion to 102.6 billion parameters. The architecture of Yuan 2.0 is designed by incorporating Attention with localized filtering, which brings a better accuracy than vanilla Attention. The proposed distributed training method with non-uniform pipeline parallel, data parallel, and optimizer parallel greatly reduces the bandwidth requirements of intra-node communication, and leads to good performance in large-scale distributed training. The Yuan 2.0 models demonstrate their good ability on code generation, math, as well as chat, compared with existing models. We plan to make progressive improvements on Yuan 2.0 in the future work.

# Appendix A: Chat examples of Yuan 2.0

## A.1 Translation
**Prompts:**
在 2011 年 10 月，Diaspora 宣布他们开始新一轮的资金筹集。

你可以帮我把这句话翻译成英文吗？(Can you help me translate the sentence into English?)

**Outputs (English Translation):**
In October 2011, Diaspora announced that they had begun a new fundraising campaign.

## A.2 Keyword Recognitiom
**Prompts:**
提出了一种基于支持向量机的滚动轴承状态监测方法。该方法采用小波能量表征轴承信号的特征，采用支持向量机作为分类器构建支持向量机状态监测系统。仿真实验表明该方法具有很好的在线监测性能和分类性能，该方法对于滚动轴承的监测具有一定的应用价值，并可方便地推广到其他类似的工业应用领域。

帮我识别出文本中的关键词：

This paper presents a method of rolling bearing condition monitoring based on support vector machine. In this method, wavelet energy is used to characterize the bearing signal, and support vector machine (SVM) is used as classifier to construct a state monitoring system. The simulation results show that the method has good on-line monitoring and classification performance. The method has certain application value for the monitoring of rolling bearings, and can be easily extended to other similar industrial applications. Help me identify key words in the text:

**Outputs (English Translation):**
支持向量机;状态监测;小波能量;滚动轴承

 (Support vector machine; Condition monitoring; Wavelet energy; Rolling bearing)

## A.3 Sentimental Analysis
**Prompts:**
情感分析：帮人买的,尽管这个人是吃饱没事干买回来做的,但是她说不错,想来应该还可以。

评论表达了哪种情感?

选项：积极，消极

Emotional analysis: Buying it for someone else, although she bought this for nothing, she said it was good and well enough. What emotion does the comment express?
Options: Positive or negative.

**Outputs (English Translation):**
积极（Positive）

## A.4 Summary
**Prompts:**
8 月 5 日起，天津航空公司将开通泉州-义乌、泉州-南宁往返航班，每周一、三、五、日执行，航班号 GS7579/80。该航班上午 8：00 南宁起飞，10：00 到达泉州，10:40 泉州起飞，12：00 到达义乌；12：40 义乌起飞，14：00 到达泉州。

使用简洁精炼的语言为以上文章写一篇摘要。摘要需要保证原文关键信息的完整度和准确性，不要随意篡改文章中的人名、地名等实体内容。

摘要：

From August 5th, Tianjin Airlines will open round-trip flights between Quanzhou and Yiwu, Quanzhou and Nanning. These flights will operate on Mondays, Wednesdays, Fridays, and Sundays with the flight number GS7579/80. The flight from Nanning departs at 8:00 AM and arrives in Quanzhou at 10:00 AM. The flight from Quanzhou departs at 10:40 AM and arrives in Yiwu at 12:00 PM. The return flight from Yiwu departs at 12:40 PM and arrives in Quanzhou at 2:00 PM.

Write a summary of the above article in concise language. It is necessary to ensure the integrity and accuracy of the key information in the original text, and do not tamper with the physical content of the article such as the name and place name.

Abstract:

**Outputs (English Translation):**

天津航空公司将于8月5日起开通泉州-义乌、泉州-南宁往返航班，每周一、三、五、日执行，航班号GS7579/80。

（Tianjin Airlines will open Quanzhou-Yiwu and Quanzhou-Nanning round-trip flights starting from August 5th. The flights will operate every Monday, Wednesday, Friday, and Sunday with the flight number GS7579/80.）

### A.5 Mails writing

**Prompts:**

我希望了解一下这个楼盘的具体情况，包括位置、户型设计、配套设施和价格等方面的信息，这样我才能做出购房决策。您能帮我生成一份邮件吗？

I would like to know the specific situation of this property, including location, unit design, supporting facilities and price information, so that I can make a purchase decision. Can you generate an email for me?

**Outputs (English Translation):**

当然可以，以下是一份您可以发送给开发商或销售代表的邮件草稿，您可以根据需要进行修改：

尊敬的销售代表，

我很感兴趣了解贵公司所代理的楼盘情况，并希望了解更多相关信息以帮助我做出明智的购房决策。

首先，请提供楼盘的具体位置和周边环境信息。我想了解该楼盘所在的城市或地区，以及周边的交通情况、生活配套和商业中心。

其次，请告知我该楼盘的户型设计。我对于房屋的空间布局和功能分区很感兴趣。请提供详细的平面图、房间尺寸以及室内装饰等方面的信息。

此外，我也很关心小区的配套设施，例如停车位、保安系统、公共区域等。请提供相关细节信息。

最后，请提供该楼盘的价格范围。我想了解该楼盘的市场价格范围，以便我做出相应的预算调整。

感谢您的帮助，期待您的回复。

此致

敬礼，

[您的姓名]

（Sure, here is a draft email that you can send to the developer or sales representative, which you can modify as needed: Dear sales representative,

I am very interested in learning more about the properties your company represents and would like to gather additional information to help me make an informed purchasing decision.

Firstly, please provide specific details regarding the location of the property and its surrounding environment. I would like to know the city or area where the property is situated, as well as information about transportation, amenities, and nearby commercial centers.

Secondly, please inform me about the layout design of the property. I am particularly interested in the spatial arrangement and functional zoning of the housing units. Please provide detailed floor plans, room dimensions, and information about interior finishes.

Additionally, I am also concerned about the facilities available within the community, such as parking spaces, security systems, and common areas. Please provide relevant details in these areas.

Lastly, kindly provide a price range for the property. I would like to understand the market price range for the development so that I can make appropriate budget adjustments.

Thank you for your assistance, and I look forward to your response.

Yours sincerely,

[Your Name]）

## A.6 Opening statement

**Prompts:**

请帮我撰写一篇企业年会的主持人开场白。

Please help me write an opening statement for the company's annual meeting

**Outputs (English Translation):**

尊敬的各位嘉宾、亲爱的同事们，大家早上好/下午好/晚上好！

我是今天晚上的主持人 XXX，非常荣幸能够站在这个舞台上，为大家带来这场盛大的年会。首先，我代表 XX 公司向各位嘉宾表示最热烈的欢迎和衷心的感谢！感谢您们从百忙之中腾出时间，与我们共度美好时光。

时光荏苒，转眼间我们又走过了一年，这一年是辛勤拼搏的一年，是团队凝聚力不断提升的一年，也是我们共同成长的一年。年会是一个回顾过去，展望未来的重要节点，同时也是我们放松心情，享受欢乐的时刻。

今天的年会，我们将为大家准备了丰富多彩的节目，有精彩的表演、有幽默的小品、还有激动人心的抽奖环节，相信每个人都能找到自己喜欢的节目和参与其中的乐趣。

同时，年会也是感恩的时刻。感谢全体员工的辛勤付出和无私奉献，感谢各位领导的悉心关怀和睿智指导，感谢合作伙伴的支持与信任。正是有了大家的团结协作和共同努力，公司才能不断发展壮大。

最后，我衷心祝愿每位同事和嘉宾在新的一年里，身体健康、工作顺利、家庭幸福！让我们携手并肩，开启新的征程，共同书写更加辉煌的明天！

感谢大家的到来，祝愿年会圆满成功！接下来，就让我们一起欢乐地度过这美好的时刻吧！

谢谢大家！

（Ladies and gentlemen, dear colleagues, good morning/afternoon/evening!

I am XXX, the host for tonight's event. It is my great honor to stand on this stage and bring you this grand annual gathering. First of all, on behalf of XX company, I would like to extend our warmest welcome and heartfelt thanks to all the distinguished guests. We appreciate your presence and taking time out of your busy schedules to join us for this wonderful occasion.

Time flies, and in the blink of an eye, another year has passed. It has been a year of hard work and dedication, a year of increasing team cohesion, and a year of collective growth. The annual gathering is an important milestone for us to review the past and look forward to the future. It is also a moment for us to relax and enjoy the joyous atmosphere.

Today's event features a diverse array of programs, including exciting performances, humorous sketches, and thrilling lucky draws. We believe that everyone will find something they enjoy and will have fun participating in the activities.

Moreover, the annual gathering is a time for gratitude. We would like to thank all the employees for their hard work and selfless dedication, appreciate the caring support and wise guidance from our leaders, and express our gratitude to our partners for their support and trust. It is through the unity, collaboration, and joint efforts of everyone that our company can continue to grow and thrive.

Finally, I sincerely wish every colleague and guest good health, successful work, and happiness in their families in the coming year! Let us join hands and embark on a new journey together to create an even more glorious future!

Thank you all for being here, and I wish the annual gathering a resounding success! Now, let's enjoy this wonderful moment together!

Thank you!)

## A.7 Work summary

**Prompts:**

我需要一篇半年护理工作总结,不知道你能不能帮我写?

I need a six-month nursing work summary, could you write it for me?

**Outputs (English Translation):**

半年护理工作总结

这半年来,我在护理领域积极承担责任,努力提供优质的医疗护理服务。以下是我在过去六个月里的工作总结:

1. 知识与技能提升:我积极参加各类进修培训,不断扩充自己的专业知识和技能。通过学习最新的护理理论和实践,我能够更好地应对各种疾病和护理问题,并为患者提供准确和及时的护理。

2. 对患者的关怀:作为一名护士,我始终将患者的需求和福祉放在首位。我与患者建立了良好的沟通和信任关系,在他们最困难的时刻给予了全力的支持和鼓励。我努力提供个性化的护理,确保每个患者都得到适当的照顾。

3. 协调与合作:在医疗团队中,我积极与其他医护人员合作,确保协调一致的护理计划和有效的医疗服务。我与医生、药剂师和其他护理人员密切合作,确保患者得到全面且高效的医护服务。

4. 病情观察与记录:我充分发挥自己的职责,认真观察患者的病情变化,并及时记录。通过详细和准确的护理记录,我能够提供给医疗团队必要的信息,以便制订适当的治疗计划和护理方案。

5. 危机处理能力:在紧急情况下,我能够冷静应对并妥善处理各种突发状况。我具备基本的急救技能和护理常识,能够迅速采取措施并为患者提供紧急援助。

6. 持续改进：我始终保持对工作的反思和学习态度，不断寻求提高护理质量的方法。我主动参与团队会议和病例讨论，并积极接受同事和上级的建议和指导，以便提升自身的专业水平和工作效率。

总而言之，这半年来，我努力提供具有同情心和专业精神的护理服务。我将继续不断学习和成长，为患者提供更好的照护。感谢团队中各位的支持与合作，相信在大家的共同努力下，我们会取得更大的进步和成就。

（Six months nurse working summary

In the past six months, I have actively taken on responsibilities in the nursing field and worked hard to provide high-quality medical care services. Here is a summary of my work over the past six months:

1. Knowledge and skills improvement: I have actively participated in various continuing education and training programs to continuously expand my professional knowledge and skills. By studying the latest nursing theories and practices, I am better equipped to handle various diseases and nursing issues and provide accurate and timely care for patients.

2. Patient care and compassion: As a nurse, I always prioritize the needs and well-being of patients. I have built good communication and trust with patients and provided full support and encouragement during their most difficult times. I strive to provide personalized care and ensure that each patient receives appropriate attention.

3. Coordination and teamwork: Within the medical team, I actively collaborate with other healthcare professionals to ensure coordinated care plans and effective medical services. I work closely with doctors, pharmacists, and other nursing staff to ensure that patients receive comprehensive and efficient healthcare services.

4. Observation and documentation of patient conditions: I fulfill my responsibilities by carefully observing changes in patients' conditions and promptly documenting them. Through detailed and accurate nursing records, I can provide the necessary information to the medical team for developing appropriate treatment plans and care strategies.

5. Crisis management skills: In emergency situations, I can calmly respond and handle various unexpected circumstances. I possess basic first aid skills and nursing knowledge, allowing me to take immediate measures and provide emergency assistance to patients.

6. Continuous improvement: I maintain a reflective attitude towards my work and a commitment to learning, constantly seeking ways to improve the quality of care. I actively participate in team meetings and case discussions, and willingly accept advice and guidance from colleagues and superiors to enhance my professional level and work efficiency.

In conclusion, over the past six months, I have been dedicated to providing compassionate and professional nursing services. I will continue to learn and grow, striving to provide better care for patients. I appreciate the support and cooperation of everyone in the team, and believe that together, we will achieve greater progress and accomplishments.）

### A.8 Chinese poetry
**Prompts:**
为我生成一篇七言绝句，主题是"相望"。
Generate a seven-sentence quad for me on the theme of "looking at each other.

**Outputs (English Translation):**
不作风流时样妆，月明宫殿舞霓裳。岂知秋雨偏相望，醉著歌声犹未凉。

（Without adornment, I dance the "Rainbow Skirt and Feathered Coat" in the moonlit palace. Little did I know that we would meet under the autumn rain, and even in my drunkenness, the music and singing have not yet cooled.）

## A.9 Chinese poetry
**Prompts:**
帮我写一首宋词，词牌名是浣溪沙，突出表现"春困"，"柳腰"，"兜鞋"。

Help me write a song poem, the name of the poem is Huan Xi Sha, highlighting the performance of "spring sleepiness", "willow waist", "pocket shoes".

**Outputs (English Translation):**
香靥凝羞一笑开。柳腰如醉暖相挨。日长春困下楼台。照水有情聊整鬓，倚阑无绪更兜鞋。眼边牵系懒归来。

（The fragrance of orchids lingers as she smiles, her slender waist warm against mine. As the day lengthens, she grows weary and leans on the terrace. The reflection in the water gently arranges her hair, while she idly mends her shoes leaning against the rail. Her eyes show a hint of attachment as she lazily returns.）

## A.10 RAG
**Prompts:**
系统软件是指控制和协调计算机及外部设备,支持应用软件开发和运行的系统，是无需用户干预的各种程序的集合，主要功能是调度，监控和维护计算机系统；负责管理计算机系统中各种独立的硬件，使得它们可以协调工作。系统软件使得计算机使用者和其他软件将计算机当作一个整体而不需要顾及到底层每个硬件是如何工作的。

系统软件是负责管理计算机系统中各种独立的硬件，使得它们可以协调工作。系统软件使得计算机使用者和其他软件将计算机当作一个整体而不需要顾及到底层每个硬件是如何工作的。一般来讲，系统软件包括操作系统和一系列基本的工具（比如编译器，数据库管理，存储器格式化，文件系统管理，用户身份验证，驱动管理，网络连接等方面的工具）。

软件系统(Software Systems)是指由系统软件、支撑软件和应用软件组成的计算机软件系统，它是计算机系统中由软件组成的部分。它包括操作系统、语言处理系统、数据库系统、分布式软件系统和人机交互系统等。

请仔细阅读上面的文章，并且回答下面的问题。

请使用文章中明确列出的信息，不要编造文章中未列出的信息，不要试图编造答案。

问题：系统软件有哪些?

System software refers to the control and coordination of computers and external equipment, support application software development and operation of the system, is a collection of various programs without user intervention, the main function is to schedule, monitor and maintain the computer system; Responsible for managing the various independent hardware in the computer system so that they can work together. System software allows computer users and other software to see the computer as a whole without having to consider how the underlying hardware works. System software is responsible for the management system of independent computer hardware, enabling them to coordinate work. System software allows computer users and other software to see the computer as a whole without having to consider how the underlying hardware works. In general, system software includes the operating system and a set of basic tools (such as compilers, database management, memory formatting, file system management, user authentication, driver management, network connectivity, etc.).

Software system (Software Systems) is to point to by the system Software, supporting Software and application Software of computer Software system, which is part of the computer system is composed of Software. It includes operating system, language processing system, database system, distributed software system and human-computer interaction system.

Please carefully read the above article, and answer the following questions.

Please use the article explicitly listed information, do not are not listed in the article information, don't try to make the answer.
Question: what are the system software?
**Outputs (English Translation):**
系统软件包括操作系统和一系列基本的工具，如编译器、数据库管理、存储器格式化、文件系统管理、用户身份验证、驱动管理、网络连接等方面的工具。

（System software includes the operating system and a series of basic tools, such as compilers, database management, memory formatting, file system management, user authentication, driver management, network connectivity, etc.）

## A.11 RAC
**Prompts:**
已知内容:

Ajax++。基本简介。传统的 Web 应用允许用户端填写表单(form)，当提交表单时就向 Web 服务器发送一个请求。服务器接收并处理传来的表单，然后送回一个新的网页。这个做法浪费了许多带宽，因为在前后两个页面中的大部分 HTML 代码往往是相同的。由于每次应用的交互都需要向服务器发送请求，应用的响应时间就依赖于服务器的响应时间。这导致了用户界面的响应比本地应用慢得多。。与此不同，AJAX 应用可以仅向服务器发送并取回必需的数据，它使用 SOAP 或其它一些基于 XML 的页面服务接口，并在客户端采用 JavaScript 处理来自服务器的响应。

Ajax++。基本简介。 AjaxAJAX 全称为"Asynchronous JavaScript and XML"（异步 JavaScript 和 XML），是一种创建交互式网页应用的网页开发技术。它使用：使用 XHTML+CSS 来表示信息；使用 JavaScript 操作 DOM（Document Object Model）进行动态显示及交互；使用 XML 和 XSLT 进行数据交换及相关操作；使用 XMLHttpRequest 对象与 Web 服务器进行异步数据交换；使用 JavaScript 将所有的东西绑定在一起。

问题:ajax 服务器端怎么写？

根据已知内容，可以找到问题对应的答案吗？只回复是或者否就行，不用回复其他内容。
Content:
Ajax++. Basic introduction. Traditional Web applications allow a client to fill out a form and send a request to the Web server when the form is submitted. The server receives and processes the incoming form and sends back a new web page. This wastes a lot of bandwidth because most of the HTML code in the two pages is often the same. Since each application interaction requires a request to be sent to the server, the application's response time depends on the server's response time. This results in a user interface that is much slower to respond than native applications. In contrast, an AJAX application can send and retrieve only the necessary data to the server, using SOAP or some other XML-based page service interface, and using JavaScript on the client side to process the response from the server. Ajax++. Basic introduction. AjaxAJAX, all Asynchronous JavaScript and XML, is a web development technique for creating interactive web applications. It uses: XHTML+CSS to represent information; Use JavaScript to manipulate DOM (Document Object Model) for dynamic display and interaction; Use XML and XSLT for data exchange and related operations; Use the XMLHttpRequest object for asynchronous data exchange with the Web server; Use JavaScript to tie everything together.
Question: how do you write ajax server?
Based on the known article, can we find the answer to the question? Reply with yes or no, don't reply with anything else.
**Outputs (English Translation):**

否（No）

### A.12 Brainstorming
**Prompts:**
假设你是人工智能工程师，如何使用聚类算法对大规模数据进行分类？
If you're an AI engineer, how do you use clustering algorithms to classify large-scale data.
**Outputs (English Translation):**
以下是我作为一名人工智能工程师使用聚类算法对大规模数据进行分类的基本步骤：

1. 数据预处理：首先将数据进行规范化和归一化处理，以便统一数据尺度和特征表达。

2. 选择聚类算法：选择合适的聚类算法，如 K-means、层次聚类、密度聚类等。

3. 确定聚类簇数：根据经验、先验和可测量性等因素，确定聚类的簇数，可以使用肘部法、轮廓系数等方法来确定。

4. 初始化聚类中心：随机选择 k 个初始聚类中心，每个聚类中心代表簇的质心点。

5. 分配数据到簇：将每个数据点分配到最近的聚类中心所属的簇，计算每个数据点与每个聚类中心之间的欧氏距离，并选择最小值作为其所属簇。

6. 更新聚类中心：重新计算每个簇的质心点，并将其设为新的聚类中心。对于每个簇，计算簇中所有数据点的平均值，即为新的聚类中心。

7. 重复步骤 5 和 6，直到收敛或达到预设停止条件为止。当聚类中心不再变化或达到了指定的迭代次数时，就可以认为算法收敛了。

8. 评估聚类结果：为了验证算法的有效性，需要使用一些评价指标，如轮廓系数、Davies-Bouldin 指数等来评估聚类结果。

9. 聚类结果的应用：一旦聚类算法完成，就可以将数据根据聚类结果进行分区，以便更好地理解和利用数据。

总之，使用聚类算法对大规模数据进行分类需要耗费大量的时间和计算资源，但可以帮助我们更好地理解和处理数据，并从中提取有用的信息。

（Here is a translation of the steps for clustering large-scale data using clustering algorithms as a AI engineer:
1. Data preprocessing: First, normalize and standardize the data to achieve uniform data scales and feature representations.
2. Select a clustering algorithm: Choose a suitable clustering algorithm, such as K-means, hierarchical clustering, density-based clustering, etc.
3. Determine the number of clusters: Based on experience, prior knowledge, and measurability, determine the number of clusters. Methods like the elbow method or silhouette coefficient can be used for determination.
4. Initialize cluster centers: Randomly select k initial cluster centers, where each center represents the centroid of a cluster.
5. Assign data to clusters: Assign each data point to the cluster whose center is closest, calculating the Euclidean distance between each data point and each cluster center and selecting the minimum as its assigned cluster.
6. Update cluster centers: Recalculate the centroids of each cluster and set them as the new cluster centers. For each cluster, calculate the average of all data points in the cluster to obtain the new cluster center.

7. Repeat steps 5 and 6 until convergence or a predetermined stopping condition is met. When the cluster centers no longer change or the specified number of iterations is reached, the algorithm is considered to have converged.
8. Evaluate clustering results: To validate the effectiveness of the algorithm, evaluation metrics such as silhouette coefficient, Davies-Bouldin index, etc., can be used to assess the clustering results.
9. Application of clustering results: Once the clustering algorithm is complete, the data can be partitioned based on the clustering results to gain better understanding and utilization of the data.
In summary, clustering large-scale data using clustering algorithms requires significant time and computational resources but can help us better understand and process data, extracting useful information from it.）

## A.13 OpenQA
**Prompts:**
你了解什么是蛇行现象吗？它是如何发生的？
What do you know about snaking phenomenon? How does it happen.
**Outputs (English Translation):**
蛇行现象是指汽车或其他交通工具在行驶过程中呈现出曲线状或摇摆状的运动。这种现象通常发生在高速运动时，特别是在悬挂系统失效或轮胎失去抓地力时。

蛇行现象的发生是由于作用在车辆和轮胎上的力不平衡所引起的。当车辆以较高速度行驶时，轮胎与地面之间的摩擦力不够提供足够的向心力，从而使得车辆产生离心力。这种离心力会导致车辆在弯道上偏离理想的直线轨迹，进而呈现出蛇行形状的运动。

此外，蛇行现象也可能是由于悬挂系统的失效导致的。当悬挂系统失效时，车辆在行驶过程中无法提供足够的支撑力，从而使得车辆产生离心力，进而导致蛇行现象。

要消除蛇行现象，可以尝试以下方法：

1. 调整悬挂系统的参数，增加减震器或调整悬挂系统的刚度，以提供更好的悬挂效果。

2. 检查和保养车辆的轮胎，确保轮胎有足够的摩擦力和抓地力。

3. 确保车辆在行驶过程中有足够的支撑力，可以通过增加悬挂系统的支撑臂、更换高性能轮胎等方法来增加支撑力。

4. 减少或避免高速行驶，以减少离心力的作用。

希望以上信息能够帮助到您。

（The snake-like phenomenon refers to the curved or swaying motion exhibited by a car or other vehicles during operation. This phenomenon typically occurs during high-speed movement, especially when there is a suspension system failure or loss of tire traction.
The occurrence of the snake-like phenomenon is caused by an imbalance of forces acting on the vehicle and tires. When the vehicle is traveling at high speeds, the friction between the tires and the road surface may not provide sufficient centripetal force, resulting in the generation of centrifugal force. This centrifugal force causes the vehicle to deviate from the ideal straight line trajectory on curves, resulting in the snake-like motion.
Additionally, the snake-like phenomenon can also be caused by suspension system failure. When the suspension system fails, the vehicle cannot provide adequate support during operation, leading to the generation of centrifugal force and resulting in the snake-like phenomenon.
To eliminate the snake-like phenomenon, you can try the following methods:
1. Adjust the parameters of the suspension system by increasing the shock absorbers or adjusting the stiffness of the suspension system to achieve better suspension performance.
2. Inspect and maintain the vehicle's tires to ensure they have sufficient friction and traction.

3. Ensure that the vehicle has sufficient support during operation by increasing the supporting arms of the suspension system or replacing high-performance tires to increase the support force.
4. Reduce or avoid high-speed driving to minimize the effect of centrifugal force.

I hope the above information is helpful to you.）

# Appendix B: A prompt example Yuan 2.0 used to generate unit tests

根据以下问题描述与示例,生成单元测试用例。

问题描述:编写一个程序,当它接收到输入的时候,会自动将输入中的所有标点符号都替换成空格,并将所有的单词转换成小写形式。最后输出处理后的字符串。

示例:

>>> clean_text("你好,世界!This is a TEST.")

"你好 世界 this is a test"

>>> clean_text("Hello World")

"hello world"

>>> clean_text("1, 2, 3, go!")

"1 2 3 go"

>>> clean_text("")

""

>>> clean_text("What's up?")

"what s up"

def clean_text(text):

# 单元测试用例:

```python
def test_clean_text():
    assert clean_text("你好,世界!This is a TEST.") == "你好 世界 this is a test"
    assert clean_text("Hello World") == "hello world"
    assert clean_text("1, 2, 3, go!") == "1 2 3 go"
    assert clean_text("") == ""
    assert clean_text("What's up?") == "what s up"
```

问题描述:给定一个由正整数组成的列表 lst 和一个正整数 k,返回列表中第 k 小的不同整数。

示例:

>>> find_kth_smallest_number([3,1,4,1,5,9,2,6,5], 3)

>>> find_kth_smallest_number([1,2,3,4,5], 4)

>>> find_kth_smallest_number([1,1,1,1], 1)

def find_kth_smallest_number(lst, k):

# 单元测试用例:

```python
def test_find_kth_smallest_number()
    assert find_kth_smallest_number([3,1,4,1,5,9,2,6,5], 3) == 3
    assert find_kth_smallest_number([1,2,3,4,5], 4) == 4
    assert find_kth_smallest_number([1,1,1,1], 1) == 1
```

问题描述:检查给定数字列表中,是否有任何两个数字之间的距离小于给定的阈值。

示例:

>>> has_close_elements([1.0, 2.0, 3.0], 0.5)

False

>>> has_close_elements([1.0, 2.8, 3.0, 4.0, 5.0, 2.0], 0.3)

True

from typing import List

def has_close_elements(numbers: List[float], threshold: float) -> bool:

# 单元测试用例:

```python
def test<sep>
```

## Appendix C: Yuan 2.0 bandwidth requirements

In this section, we analyze the communication bandwidth requirements for the Yuan 2.0 training process.

Firstly, we estimate that the performance of Yuan 1.0 model trained on a single card is 55.4% of the theoretical peak performance of the GPU, then

The calculation time is:

$$T_0 = \frac{96ABLSH^2 \left(1 + \frac{S}{6H}\right)}{P_s * T_s * F} = 36.68$$

The idle time of pipeline bubble can be calculated with the following equation:

$$T_1 = \frac{96ABLSH^2 \left(1 + \frac{S}{6H}\right)}{P_s * T_s * F} * \frac{(P_s - 1)}{A} = 2.83$$

The time consumption of pipelined parallel communication is:

$$T_2 = \frac{8ABSH}{n_{net} * BW_{net}} = 1.31$$

The time consumption of tensor parallel communication is:

$$T_3 = 48 * \frac{L * (T_s - 1)}{P_s * T_s * BW_{link}} ABSH = 2.60$$

The time consumption of data parallel communication is:

$$T_4 = 12LH^2 \left(1 + \frac{13}{12H} + \frac{V}{12LH}\right) * \frac{8 * (D_s - 1)/D_s}{P_s * n_{net} * BW_{net}} = 0.90$$

Finally, we get the total time consumption as $T = T_0 + T_1 + T_2 + T_3 + T_4 = 44.33$ sec.

We assume that when training Yuan 2.0, the proportion of the time costs of tensor parallelism and pipeline parallelism are the same as those in training Yuan 1.0.

According to Eq (1) ~ Eq (4), the intra-bandwidth and inter-bandwidth requirement for training Yuan 2.0 102B are 730 GB/s and 43 GB/s, respectively.

**Appendix D: Hyper-Parameters to compute time consumption for each iteration of Method 1 and Method 2**

| Hyper-params | Value |
|---|---|
| Accumulate time | 384 |
| Number of Layers | 84 |
| Hidden size | 8192 |
| Sequence length | 4096 |
| Pipeline parallel size | 32 |
| Tensor parallel size | 1 |
| Data parallel size | 3 |
| Intra-node bandwidth | 85% of P2P peak bandwidth |
| Inter-node bandwidth | 2*195 Gb/s |